\documentclass[conference]{IEEEtran}
\IEEEoverridecommandlockouts

\usepackage{cite}
\usepackage{amsmath,amssymb,amsfonts}
\usepackage{algorithm}
\usepackage{algorithmic}
\usepackage{graphicx}
\usepackage{booktabs}
\usepackage{textcomp}
\usepackage{xcolor}
\usepackage{comment}

\def\BibTeX{{\rm B\kern-.05em{\sc i\kern-.025em b}\kern-.08em
    T\kern-.1667em\lower.7ex\hbox{E}\kern-.125emX}}

\begin{document}

\bstctlcite{IEEEexample:BSTcontrol}

\title{ORION: An Outcome-Driven LLM Advisor for RL-Based Multi-Domain 6G Slice Placement}

\author{\IEEEauthorblockN{Mahmoud S. Mayaleh\textsuperscript{*}}
\IEEEauthorblockA{\textit{SAMOVAR, Télécom SudParis} \\
\textit{Institut Polytechnique de Paris}\\
Palaiseau, France \\
mayaleh@lix.polytechnique.fr}
\and
\IEEEauthorblockN{Fetia Bannour}
\IEEEauthorblockA{\textit{ensIIE, SAMOVAR} \\
Évry, France \\
fetia.bannour@telecom-sudparis.eu}
\and
\IEEEauthorblockN{Nour-El-Houda Yellas}
\IEEEauthorblockA{\textit{SAMOVAR, Télécom SudParis} \\
\textit{Institut Polytechnique de Paris}\\
Palaiseau, France \\
nour-el-houda.yellas@telecom-sudparis.eu}
}

\maketitle

\begingroup
\renewcommand\thefootnote{}
\footnotetext{\textsuperscript{*}Work done while at SAMOVAR, T\'el\'ecom SudParis. Now with LIX, \'Ecole polytechnique, Institut Polytechnique de Paris.\\
Accepted at IEEE MobiWac 2026.\\
\copyright~2026 IEEE. Personal use of this material is permitted. Permission from IEEE must be obtained for all other uses, in any current or future media, including reprinting/republishing this material for advertising or promotional purposes, creating new collective works, for resale or redistribution to servers or lists, or reuse of any copyrighted component of this work in other works.}
\addtocounter{footnote}{-1}
\endgroup

\begin{abstract}
Deploying 6G network slices across administrative domains requires an online decision for each arriving request: which domain hosts each virtual network function, how
traffic crosses domain boundaries, and whether the slice is admitted. This joint decision problem is NP-hard under continuous arrivals. Experiments on a parametrized family of multi-domain topologies show that no fixed planning strategy dominates across all regimes: the performance ranking reverses between topology classes, and feasible placements exist for most rejected requests but remain undiscovered. These findings identify the plan layer as the decisive performance lever. We introduce ORION, a hierarchical orchestrator in which two LLM agents translate service intents into structured requests and generate abstract placement plans. A cooperative multi-agent RL
coordinator selects the cross-domain partition, treating the plan as a persistent soft prior, while per-domain agents place the functions. ORION main contribution lies at the plan
layer: an outcome-driven episodic memory that records, for each completed attempt, the network-condition profile, the proposed plan, and its verified feasibility outcome. The memory feeds verified successes and labelled failures back to the planner, thereby closing the adaptation loop using deployment feasibility rather than a scalar reward assigned to the planner's own proposals. To isolate the effect of the plan quality, we evaluate the planner on held-out topology families while keeping the downstream selection and execution mechanisms fixed. Under this controlled setting, the proposed memory improves admission by $4.6$--$11.5$ percentage points over the same planner without memory, while importance-based retention accounts for most of the improvement over a first-in-first-out store.
\end{abstract}


\section{Introduction}

Network slicing enables a shared 6G physical infrastructure to be partitioned into multiple logical networks tailored to different service requirements~\cite{ebrahimi2024}. Network
function virtualization makes this possible: the functions that once ran on dedicated middleboxes are deployed as virtual network functions (VNFs) chained into a service function chain (SFC) and instantiated on commodity compute. In the multi-provider 6G setting considered in this work, the physical substrate spans multiple administrative domains rather than a single operator infrastructure. Each domain controls its own resource capacity and a subset of compute tiers---including RAN and edge sites, regional clouds, and central clouds---while inter-domain links are more constrained than intra-domain connectivity. Deploying a slice therefore requires determining which domain hosts each VNF in its service chain, how inter-VNF traffic is routed across domain boundaries, and whether the slice request should be admitted.

These decisions must be automated. Slice requests arrive continuously and must be resolved online; each domain exposes aggregate state rather than per-node inventory across its management boundary, so the orchestrator plans without a global per-node view. This separation holds even under a single operator: a compact per-domain interface is what keeps orchestration scalable and domains independently operable.
The QoS contract couples placement to routing, since a chain split across domains pays inter-domain propagation delay and consumes transit bandwidth that a co-located chain does not.
The resulting joint placement, routing, and admission problem is NP-hard, generalizing both bin packing \cite{garey1979computers} and integer multi-commodity flow
\cite{even1975complexity}. Exact formulations \cite{addis2015} are intractable online, whereas heuristics are fast but encode a fixed strategy that does not adapt when the substrate, workload mix, or load regime changes.

Two lines of work address this challenge. Learning-based orchestrators replace fixed placement strategies with policies trained through interaction with the substrate~\cite{flagvne}, but the resulting policies remain tied to the topology distribution seen during the training. Large language models have entered network management as intent translators~\cite{llm-net-dandoush,llm-net-nam,llm-net-su}, and more recently as
advisors that bias or guide a learned controller
\cite{linvit-hu,yan-priors,saycan,oranguide-lotfi,lotfi-prompttuned}. LLMs are well suited to this setting because they can reason over the abstract per-domain information that operators are willing to expose. At this level, identifying which tier assignments and cross domain partitions remain feasible for a given substrate structure relies more on operator expertise than on numerical optimization. However, existing approaches typically keep the LLM fixed after deployment. A smaller body of work allows the LLM to adapt online, but only through scalar performance feedback on its own
proposals~\cite{arc-shokrnezhad,wang-agentic-slicing}. To our knowledge, none of these approaches conditions this adaptation on a verified feasibility outcome associated with a concrete plan and the network conditions under which it was evaluated.

We present ORION, a hierarchical orchestrator for multi-domain slice placement. Two LLM agents respectively translate the service intent into a structured request and propose an abstract placement plan. A multi-agent RL coordinator converts this proposal into a concrete cross-domain partition, treating it as a soft prior, while per-domain agents place VNFs and route their flows.
ORION main contribution lies at the plan layer: an outcome-driven episodic memory records the network-condition profile, the proposed plan, and the verified feasibility outcome for each completed attempt, and feeds verified successes and labelled failures back to the planner. A
diagnostic study on a parametrized family of topology identifies the plan layer as the effective point of intervention and motivates this design.

This paper makes the following contributions.
\begin{itemize}
    \item A verifier-based diagnostic study showing that planning failures primarily arise from search and routing limitations and that no fixed plan builder dominates across topology regimes.
    \item Orion, a hierarchical architecture  combining LLM-based planning, RL-based cross-domain partitioning under a soft-prior, and per-domain node placement.
    \item An outcome-driven episodic memory that provides the planner with verified successes and labelled failures retrieved under similar network conditions.
    \item  A controlled evaluation on held-out topology families that isolates the effect of memory against reward-labelled control and measures the LLM invocation overhead.
\end{itemize}

The rest of the paper is organized as follows. Section~\ref{sec:related} reviews related
work. Section~\ref{sec:model} states the
system model and the VNF placement problem.
Section~\ref{sec:diagnosis} presents the diagnostic
characterization that locates the lever at the plan layer.
Section~\ref{sec:arch} describes the architecture, the LLM agents and their memory protocol. Section~\ref{sec:rl} presents the RL selector. Section~\ref{sec:exp} reports the experimental setup and results, and
Section~\ref{sec:conclusion} concludes the paper.

\section{Related Work}
\label{sec:related}

Multi-domain slice placement has been studied primarily as an optimization problem: given an SFC, a substrate partitioned into administrative domains, and a
QoS contract, decide the VNF-to-domain assignment, the inter-domain routing, and
the admission verdict. Exact formulations \cite{addis2015} and the heuristics that approximate them online share the property that the placement strategy is fixed in
advance; learned orchestrators \cite{bannour2023} make it adaptive but tie it to the substrate it was
trained on. The recent turn to language models cuts across this axis, and it is that axis we use to organize what follows.

\subsection{LLMs as priors for reinforcement learning}
ORION couples its planner to the RL selector through a persistent soft prior, a mechanism we inherit from the value-based view of LLM-guided RL.~\cite{linvit-hu} formulates the LLM as a prior over actions with formal guarantees and use it to steer value-based learning (LINVIT). \cite{yan-priors} generalizes the LLM-as-prior idea
through a variational objective, and \cite{saycan} is recovered
as the greedy limit of the same KL-regularized objective, grounding LLM proposals in a learned affordance value. In all three the LLM is fixed: it shapes the policy but
never updates from what the policy does.

\subsection{LLM-augmented RL for network slicing}
The closest line of work is the LLM-augmented DRL approach to O-RAN slicing in \cite{lotfi-icc2025}, a RAG-driven prompt-learning variant (ORAN-GUIDE) in \cite{oranguide-lotfi}, and a
prompt-tuned variant for dynamic slicing in \cite{lotfi-prompttuned}. Across these approaches, the LLM is
trained or tuned offline and held fixed at deployment and does not report online adaptation based on the outcomes of executed controller decision. A second body of work adapts the LLM conditioning during deployment, making the distinction from ORION more subtle.
For instance,~\cite{arc-shokrnezhad} updates reasoning exemplars with realized action rewards and retrieve high- and low-reward cases into the LLM few-shot prompt to guide high-level action sequencing. \cite{wang-agentic-slicing} describes an agentic-AI framework for QoE-aware slicing in Industrial IoT whose incremental memory logs observed performance and folds it back into a retrieval-augmented intent-inference stage. Both approaches therefore exploit deployment feedback, but at different stages:~\cite{arc-shokrnezhad} associates scalar rewards with action sequences, whereas~\cite{wang-agentic-slicing} uses performance and QoE observations to refine intent inference. Neither conditions the LLM-based component on a verifier-checked feasibility outcome tied to a concrete plan structure and the network conditions under which it was evaluated, which is the signal stored by ORION episodic memory.

\subsection{Agentic memory}
ORION's memory follows the cognitive-architecture framing of CoALA \cite{coala-sumers},
which separates semantic, episodic, and procedural memory for language agents; our split
between read-only operator knowledge ($K^B$) and read-write episodic outcomes ($M^B$) is
a direct instance. General-purpose agentic-memory systems such as A-MEM \cite{amem-xu}
and Mem0 \cite{mem0-chhikara} supply lifecycle and extract-update pipelines for LLM-agent
memory, but over conversational or task-completion traces. ORION differs in what is stored
and to what end: entries are keyed on topology signature and record the plan shape
together with its ground-truth feasibility outcome, so retrieval steers the planner toward
strategies that verifiably survived under similar substrate conditions rather than toward
dialogue or task-completion history.

\subsection{Plan reuse and call economy}
ORION limits LLM inference overhead through a plan cache indexed by request signatures. Cached plans are revalidated before reuse and invalidated when the underlying conditions drift. We position this against Agentic Plan Caching \cite{apc2025}, which reuses plan templates at test time to cut LLM calls. Relative to it, our cache adds feasibility revalidation and drift-triggered invalidation on reuse; it is validated engineering for call economy, orthogonal to the outcome-loop contribution, and not
claimed as novelty.

\begin{table}[t]
  \centering
  \caption{Comparison of LLM-assisted orchestration approaches.\textbf{G}: guides an RL controller; \textbf{A}: adapts from deployment outcomes; \textbf{C}: uses verifier-checked plan feedback.}
  \label{tab:related}
  \small
  \setlength{\tabcolsep}{4pt}
  \begin{tabular}{@{}p{0.66\columnwidth}ccc@{}}
    \toprule
    \textbf{Mode} & \textbf{G} & \textbf{A} & \textbf{C} \\
    \midrule
    Intent translator only \footnotesize(most slicing work)
      & $\times$ & $\times$ & $\times$ \\[2pt]
    Static LLM prior/advisor to RL \footnotesize(\cite{linvit-hu}; \cite{yan-priors}; \cite{saycan}; ORAN-\cite{oranguide-lotfi}; \cite{lotfi-icc2025})
      & \checkmark & $\times$ & $\times$ \\[2pt]
    Adapts LLM from scalar reward on own proposals / aggregate perf.\ \footnotesize(\cite{arc-shokrnezhad}; \cite{wang-agentic-slicing})
      & \checkmark & \checkmark & $\times$ \\
    \midrule
    \textbf{ORION (our work)}
      & \checkmark & \checkmark & \textbf{\checkmark} \\
    \bottomrule
  \end{tabular}
\end{table}
Table~\ref{tab:related} summarizes the main differences among the cited approaches.

\section{System Model and Problem Formulation}
\label{sec:model}

We orchestrate service function chains over a multi-domain substrate under admission control. This section defines the substrate and request models and presents a compact MILP formulation of the joint admission, placement and routing problem. ORION addresses these decisions online under dynamic request arrival.

\subsection{Multi-domain substrate}

We consider a multi-provider cloud-edge infrastructure in which an MNO orchestrates slice functions across its own resources and capacity leased from third-party infrastructure providers. The infrastructure is partitioned into $M$ administrative domains, each operated by a provider and associated with an infrastructure tier $\tau \in \{\text{RAN-edge},\text{MEC},\text{regional-cloud},\text{central-cloud}\}$.
A provider owns an administrative domain and retains control of its node-level infrastructure and exposes only aggregate resource and SLA information to the MNO.

Domain $m \in M$ is represented as an undirected subgraph $\mathcal{G}^m=(\mathcal{N}^m,\mathcal{L}^m)$ comprising compute
nodes and intra-domain links. Domains are joined through inter-domain links $\mathcal{L}^{\mathrm{inter}}$, forming the undirected weighted substrate
$G^P=(\mathcal{N},\mathcal{L})$ where
[
$\mathcal{N}=\bigcup_m\mathcal{N}^m$,
\qquad
$\mathcal{L}=\bigcup_m\mathcal{L}^m\cup\mathcal{L}^{\mathrm{inter}}$. 
]
A node $n$ is characterized by its CPU capacity $C_n$, RAM capacity $R_n$, infrastructure tier $\tau_n$, domain membership $m(n)$, and a profile processing delay $\delta_{f,n}\!\ge\!0$ for VNF $f$. Each link $\ell$ has a bandwidth capacity $B_\ell$, propagation delay $D_\ell$, and type
$\theta_\ell\in\{\mathrm{intra},\mathrm{inter}\}$; inter-domain links are assigned a higher routing-cost coefficient. The MILP treats $\delta_{f,n}$ as a fixed profiled baseline, where as the
online solution applies an $M/M/1$ sojourn-time model to capture the increase in processing delays with node utilization.

\subsection{Network slice requests}
We represent a slice request by a tuple $s=(\mathcal{F}_s,\mathcal{E}_s,\mathbf{q}_s,\mathcal{D}_s)$ where $\mathcal{F}_s=\{f_1^s\ldots,f_{K_s}^s\}$ is an ordered VNF chain, $\mathcal{E}s={(f_k^s,f_{k+1}^s)\mid 1\leq k<K_s}$ is its set of directed flow edges, $\mathbf{q}_s=(D^{\max}_s,\beta^{\mathrm{in}}_s)$ specifies the end-to-end delay budget and ingress data rate, and $\mathcal{D}_s$ defines the eligible nodes for each VNF according to tier and domain constraints. Each VNF $f_k^s$ requires minimum CPU and RAM resources, denoted by $\tilde{c}_{f_k^s}$ and $\tilde{r}_{f_k^s}$ respectively. %  demand $\tilde{c}_{f_k^s}$, a minimum RAM demand $\tilde{r}_{f_k^s}$, 
The volume change ratio (VCR) $\rho_{f_k^s}\in\mathbb{R}_+$ scales the flow it processes ($\rho<1$
compression, $\rho\!\approx\!1$ pass-through, $\rho>1$ expansion). The VCR sets the
bandwidth required on each downstream flow edge:
\begin{equation}\label{eq:vcr-bw}
  \beta^s_{k,k+1}=\beta^{\mathrm{in}}_s\textstyle\prod_{j=1}^{k}\rho_{f_j^s},
\end{equation}
This formulation accounts for traffic transformations along the chain, whereas a uniform per-edge rate could incorrectly dimension links downstream of volume-changing VNFs.

The request model combines standardized service-level attributes, such as latency and throughput requirements, with the orchestration-specific information required for placement. The former are aligned with the GSMA Generic Network Slice Template~\cite{gsma-ng116}, while the VNF chain, per-function resource demands, and placement eligibility are orchestration-specific extensions.

\subsection{Mathematical modeling}
For each slice we decide binary placement variables $x_{f_k^s,n}$, CPU and RAM allocation $a^c_{f_k^s,n},a^r_{f_k^s,n}\ge0$, binary routing variables $y_{\ell,f_k^s,f_{k+1}^s}$, per-flow bandwidth allocation $a^b_{\ell,f_k^s,f_{k+1}^s}\geq0$, and the admission variable $z_s\in\{0,1\}$. Here, $y_{\ell,f_k^s,f_{k+1}^s}=1$ indicates that physical link $\ell$ carries the flow from $f_k^s$ to $f_{k+1}^s$, according to the order of the service chain. 

An admitted slice must satisfy: \textbf{C1}
placement uniqueness ($\sum_n x_{f_k^s,n}=z_s$); \textbf{C2}/\textbf{C3} node CPU and RAM capacities; \textbf{C4} resource allocation constraints, which link CPU and RAM allocations to placement and require each hosted VNF to receive at least its minimum demands; \textbf{C5} link bandwidth capacity; \textbf{C6} flow conservation, such that the links selected by $y$ variables form a valid path between the hosts of consecutive VNFs; \textbf{C7} the end-to-end delay budget; and \textbf{C8} placement eligibility ($x_{f_k^s,n}=0$ for $n\notin\mathcal{D}_{f_k^s}$). Three constraints that are particularly relevant to the considered scenarios are stated below. Constraints~\textbf{C5b} allocates at least the required flow rate $\beta^s_{k,k+1}$ to each link used by the flow, while assigning no bandwidth to unused links and respecting the link capacity $B_\ell$:
\begin{equation}\label{eq:c5b}
  \beta^s_{k,k+1} y_{\ell,f_k^s,f_{k+1}^s}   \leq a^b_{\ell,f_k^s,f_{k+1}^s}\,
  \leq B_\ell y_{\ell,f_k^s,f_{k+1}^s}
  \quad
  \substack{\forall \ell\in\mathcal{L},\\
  \forall(f_k^s,f_{k+1}^s)\in\mathcal{E}_s}.
  \tag{C5b}
\end{equation}

The end-to-end delay budget combines VNF processing and routed propagation delays:
\begin{equation}\label{eq:c7}
  \textstyle\sum_{k}\!\Big(\sum_{\ell}D_\ell y_{\ell,f_k^s,f_{k+1}^s}
  +\sum_{n}\delta_{f_k^s,n}x_{f_k^s,n}\Big)\le D^{\max}_s z_s;
  \tag{C7}
\end{equation}
An optional inter-domain hop limit bounds the number of domain-crossing links used by each inter-VNF flow:
\begin{equation}
\sum_{\ell\in\mathcal{L}^{\mathrm{inter}}}y_{\ell,f_k^s,f_{k+1}^s} \leq H,
\forall(f_k^s,f_{k+1}^s)\in\mathcal{E}_s.
\tag{C9}
\end{equation}
The objective jointly maximizes slice admission while minimizing CPU, RAM and routing costs. Because each admitted VNF receives its requested CPU and RAM, the goal is to favor less resource-intensive sets of slices when admission decisions trade off. 
\begin{equation}\label{eq:obj}
\begin{aligned}
  \max\;\;\mu\!\sum_{s}z_s
  &-\alpha\!\!\sum_{s,f_k^s,n}\!\!(a^c_{f_k^s,n}\!+\!a^r_{f_k^s,n})\\
  &-\!\!\sum_{\ell\in\mathcal{L}^{\mathrm{intra}}}\!\!\!\!\gamma^{\mathrm{intra}}a^b_{\ell,f_k^s,f_{k+1}^s}
  \;-\!\!\sum_{\ell\in\mathcal{L}^{\mathrm{inter}}}\!\!\!\!\gamma^{\mathrm{inter}}a^b_{\ell,f_k^s,f_{k+1}^s}
\end{aligned}
\end{equation}
The weights are selected to prioritize admission over resource efficiency, while $\gamma^{\mathrm{inter}}>\gamma^{\mathrm{intra}}$ penalizes inter-domain traffic more heavily. The resulting MILP combines capacitated VNF placement and flow routing and is NP-hard, with $O(|\mathcal{F}||\mathcal{N}|+|\mathcal{L}||\mathcal{E}|)$ binary variables per slice. This complexity motivates ORION, which makes online decisions without solving the formulation at each arrival.

\subsection{Online arrival model}
Requests arrive according to a Poisson process of rate $\lambda$. Each admitted slice retains its allocated resources for an exponentially distributed lifetime with rate $\nu$ before releasing them. ORION decides admission, placement, and routing upon each arrival, while the exhaustive verifier described in Section~\ref{sec:diagnosis} provides the feasibility verdicts used for diagnosis and evaluation.

\section{Diagnosis: Locating the Lever}
\label{sec:diagnosis}

Before presenting ORION, we identify where additional intelligence most affects admission. We use a parametrized multi-domain topology generator that varies domain capacity, tier coverage and inter-domain bandwidth. An exhaustive feasibility verifier enumerates placements under the complete constraints set, including load dependent $M/M/1$ delays, VCR-dependent flow rates, and inter-domain hop limits. It provides both a per-family admission ceiling and a feasibility verdict for each rejected request. 

\subsection{Rejected slices remain feasible}
On a five-domain, tier-scarce substrate, first-fit-decreasing (FFD) rejects $26.1\%$ of arrivals across five seeds of 200 requests. % slices (five seeds, 200 arrivals each).
The verifier finds a feasible placement for every rejected request: 84.7\% are search failures, where the heuristic misses an existing placement, while the remaining $15.3\%$ require topology-aware cross domain cuts and routing choices. The primary limitation is therefore plan discovery rather than aggregate capacity.

\subsection{No fixed plan-builder dominates}
A co-location-first strategy recovers $88.9\%$ of FFD's rejections on tier-friendly topologies. On tier-hostile topologies, however, it rejects $13.8\%$ of arrivals accepted by FFD, producing a 12--13 percentage-point reversal driven solely by topology. Even a combined co-location-first strategy with FFD fallback remains, on average, 18.7 points below the verifier ceiling on bandwidth-scarce families, while the gap falls below one point when bandwidth is abundant. Hence, no fixed plan building strategy performs consistently across topology regimes. 

\subsection{The partition selector is not the main lever}
For isolated requests, optimal partition enumeration outperforms random partitioning by 23.4 percentage points. Under sequential arrivals, however, the advantage decreases to 4.5 points because locally optimal decisions deplete preferred domains and reduce future admission opportunities. Similarly, rejecting expensive arrivals with perfect knowledge of their costs and lifetimes reduces admission at every tested rejection level. CPU utilization remains below $50\%$ even at $3.2\times$ overload, confirming that aggregate compute is not the binding resource in these experiments.

Together, these results locate the main opportunity for admission improvement at the plan layer, which determines tier assignments, cross-domain cuts, and the routing structure exposed to the downstream selector.

\section{ORION architecture and LLM planning}
\label{sec:arch}

\begin{figure}[t]
  \centering
  \includegraphics[width=\columnwidth]{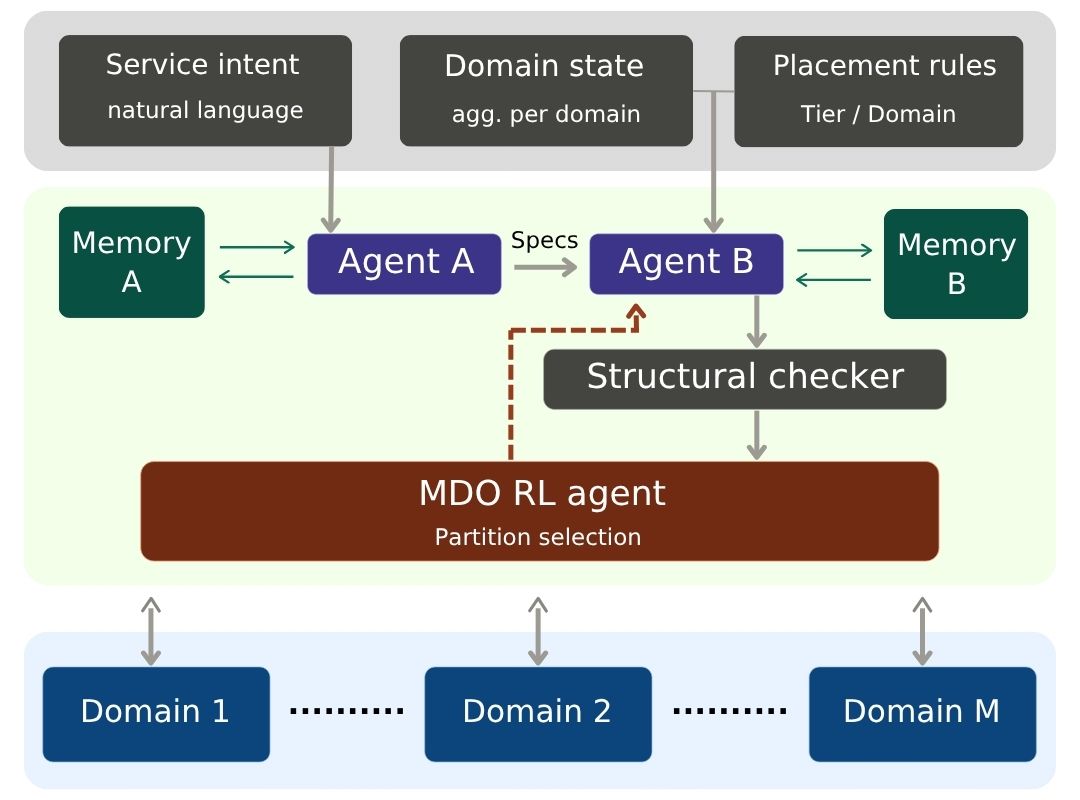}
  \caption{ORION's three-layer architecture: LLM agents propose an abstract placement, the MDO commits the cross-domain partition, and domain actors perfom node-level placement. Verified outcomes are fed back into episodic memory~$M^B$.}
  \label{fig:arch}
\end{figure}

ORION processes each arriving request through three hierarchical layers, as illustrated in Fig.~\ref{fig:arch}. The LLM translates the service intent and proposes an abstract placement plan; the reinforcement learning-based multi-domain orchestrator (MDO) commits the function-to-domain assignment and admission decision; and per-domain actors place the assigned VNFs on concrete nodes.

\subsection{Inter-domain placement}
Each arriving request is first turned into a structured slice by an LLM intent
translator (Agent~A, Section~\ref{sec:llm}). An LLM \emph{plan} layer (Agent~B,
Section~\ref{sec:llm}) then reasons at the service and tier level over an
\emph{abstract} topology (per-domain aggregate residuals and inter-domain link
summaries) and emits, for each VNF, a required tier and a suggested domain.
Crucially, the LLM never sees per-node substrate state: domain residuals are summed
over nodes, so single-node fitness is hidden from it by construction. This mirrors
what a domain operator is willing to publish across an administrative boundary.

An RL \emph{selection} layer (the MDO coordinator, Section~\ref{sec:rl}) consumes
the plan and assigns each VNF to a domain, learning where to deviate from the
suggested partition; it observes per-domain residuals, the maximum single-node
headroom per domain (which restores the node-level fitness signal the planner is
denied), and the plan's required tiers as a hard feasibility mask. The coordinator
also routes the inter-VNF flows that cross domain boundaries and owns the admission
decision. Here the abstraction is a property of the \emph{policy input}, not of the
routing mechanism: the coordinator \emph{selects} partitions from the abstract
inter-domain link summaries above, but at commit time it \emph{routes} those flows
and debits bandwidth on the full per-node substrate graph, since a concrete feasible
path must exist for the admission verdict to be sound. What the learned policy never
observes is per-node inventory; what the routing subroutine traverses is the concrete
graph.

A per-domain \emph{execution} layer (the domain actors, Section~\ref{sec:rl-actors})
then receives the fragment of the plan assigned to its domain, places those VNFs on
concrete nodes against per-node state no other layer observes, and reports back
feasibility, intra-domain delay, and cost.

\textbf{Relation to standardized management planes.} The coordinator occupies the
cross-domain management position and each domain actor a per-domain one, the roles
informally called NSMF and NSSMF \cite{3gpp-tr28801}, which 3GPP's architecture
framework places as an end-to-end management domain consuming services from
per-domain ones \cite{3gpp-ts28533}. Aggregate-only exposure across that boundary is
the model those specifications assume, since a slice may be delivered with graded
exposure of its internal structure or none at all \cite{3gpp-ts28530}, and the same separation appears in ETSI NFV MANO, where per-domain resource
control sits beneath an orchestration layer that does not address individual nodes
\cite{etsi-nfv-man001}.

Stated as an interface, the cross-domain protocol is small enough to write out in
full. \emph{Downward}, per arrival, a domain receives only its plan fragment: for
each VNF assigned to it, the required tier and the CPU, RAM, and ingress-bandwidth
demands, plus the bandwidth of any flow edge crossing its boundary. \emph{Upward},
it returns a single triple $(z^m,\delta^m,\mathrm{cost}^m)$: a feasibility flag, the
intra-domain delay it can offer, and the resource cost it would incur. Nothing else
crosses. A domain never publishes node identities, per-node residual capacity, or its
internal topology, and the signalling is $O(K)$ values per domain per arrival for a
chain of $K$ VNFs, independent of the number of nodes it operates.

\subsection{Per-arrival orchestration}
Algorithm~\ref{alg:orion} gives the end-to-end per-arrival pipeline: intent
translation, constrained planning, structural validation, partition selection under
the tier-feasibility mask, per-domain placement with inter-domain routing, single-shot
commit, and the ground-truth verification that labels the outcome back into memory.
The pre-commit check uses static per-hop delays while the verifier re-checks capacity,
bandwidth, throughput, and end-to-end delay under load-dependent M/M/1 queueing; the
residual between them is exactly the signal the outcome memory is designed to capture.

\begin{algorithm}[t]
\caption{ORION per-arrival orchestration}
\label{alg:orion}
\footnotesize
\begin{algorithmic}[1]
\REQUIRE intent, substrate $G^P$, memories $K^A,M^A,K^B,M^B$
\ENSURE admission $z_s$, committed partition $m$
\STATE $\mathrm{spec} \gets \textsc{AgentA}(\mathrm{intent}, K^A, M^A)$; $\hat G \gets \textsc{Aggregate}(G^P)$
\STATE $\tilde m \gets \textsc{AgentB}(\mathrm{spec}, \hat G, K^B, M^B)$ \COMMENT{grammar-constrained}
\IF{$\neg\,\textsc{StructuralCheck}(\tilde m)$} \RETURN reject \ENDIF
\STATE $o \gets \textsc{Observe}(G^P, \mathrm{spec})$ \COMMENT{residuals, $h^m$, tier mask}
\STATE $m \sim \pi^{\mathrm{MDO}}_\phi(\cdot \mid o)$ \COMMENT{joint partition, KL prior to $\tilde m$}
\FORALL{domains $d$ touched by $m$}
    \STATE $(z^d,\delta^d,\mathrm{cost}^d) \gets \textsc{Actor}_d(m|_d)$ \COMMENT{GATv2}
\ENDFOR
\STATE route inter-domain flows of $m$ on the full graph $G^P$
\IF{$\bigwedge_d z^d$ and \textsc{PreCommit} passes}
    \STATE commit $m$; $v \gets \textsc{Verify}(m)$ \COMMENT{M/M/1 ground truth}
    \STATE $z_s \gets 1$ if $v$ feasible, else deallocate and $z_s \gets 0$
\ELSE \STATE $z_s \gets 0$
\ENDIF
\STATE $R \gets \textsc{Reward}(z_s, \mathrm{cost}, v, \mathrm{LocalScore})$; \textsc{Write}($M^B$) on signal
\RETURN $z_s,\; m$
\end{algorithmic}
\end{algorithm}

\subsection{Intent translation and Plan generation}
\label{sec:llm}

ORION uses two LLM agents in sequence, each with its own CoALA-style memory; the
evaluation here concerns the plan layer, and Agent~A's translation study is reported
in the extended version. Agent~A
translates a natural-language service intent into a structured slice request; Agent~B
turns that request into an abstract placement plan. Both adapt through memory rather than
through weight updates; the headline contribution is Agent~B's outcome-driven memory
$M^B$.

\paragraph{Agent A: intent translation}
Agent~A maps a natural-language service intent to a structured slice request: a JSON
object giving the slice type (one of eMBB, URLLC, mMTC, V2X, XR), the VNF chain with
per-VNF resource demands and volume-change ratios, the inter-VNF flow edges, and the QoS
requirements. Translation is a single LLM call, conditioned on operator reference
knowledge and past-translation examples drawn from Agent~A's own semantic store $K^A$ and
episodic store $M^A$; the output is checked against the request schema and regenerated
under a bounded retry budget on a schema violation.

\paragraph{Agent B: constrained plan generation}
Agent~B is a frozen, instruction-tuned LLM invoked once per arrival. Its context is
assembled from four sources: the
slice request (per-VNF resource demands, permitted tiers, per-flow bandwidth and QoS
requirements), the abstract topology of Section~\ref{sec:arch}, reference entries
retrieved from the semantic store $K^B$, and labelled examples retrieved from the
episodic store $M^B$. It emits a JSON plan giving, for each VNF, a domain assignment and
a required tier, and, for each flow, a minimum bandwidth and a cross-domain-boundary flag.

Two mechanisms keep the plan measurable rather than free-form. First, decoding is
grammar-constrained: each VNF's domain field is restricted at sampling time to the
domains whose tier support makes the VNF placeable, so the model cannot emit a
tier-infeasible assignment. Second, ORION deliberately does not trust the LLM on the two
fields where it can be silently wrong: the required tier and the cross-domain-boundary
flag are recomputed from the substrate and overwrite whatever the model emitted, so they
are advisory. Plan quality is therefore carried by the domain assignments and their
induced cut points alone. A structural checker then validates the plan against aggregate
per-domain capacity, tier eligibility, and inter-domain reachability before it reaches the
selector; a structurally invalid plan is regenerated under a bounded budget, and, failing
that, the slice is rejected.

\subsection{Semantic memory $K^B$}
$K^B$ is a read-only store of operator-authored infrastructure knowledge, curated offline
and frozen before any experiment. Each entry pairs a topic and slice-type/topology tags
with a free-text body; entries cover tier conventions, inter-domain link characteristics,
known-good partitioning patterns, anti-patterns, and domain-specific properties, that is,
the expertise a human operator would encode without access to deployment outcomes.
Retrieval filters by slice-type and topology tag and ranks by lexical overlap, and the
selected entries enter Agent~B's prompt as a reference block. $K^B$ is identical across
all LLM arms, so it is the fixed knowledge baseline the episodic store is measured against.

\subsection{Episodic memory $M^B$}
$M^B$ is the outcome-driven store that makes the advisor adapt, and it is the component
the experiments isolate. Unlike $K^B$, it is written during operation. Each record pairs a
topology signature and slice features with the plan shape Agent~B produced (its co-locate
or split strategy, per-VNF tier assignment, cross-domain cut points, and the inter-domain
links used) and the plan's ground-truth outcome: whether it survived feasibility
verification, tagged by the dominant violation when it did not. Throughput-floor and
end-to-end-delay violations are first-class tags, so topology-specific failure modes can
be retrieved. The outcome is what makes the store outcome-driven: it gates which episodes
are written and how they are labelled, while the topology signature drives what is
retrieved. $M^B$ is governed by four operations.

\textbf{Write.} An episode is recorded only when it carries learning signal, judged by its
verified outcome: any constraint violation, a high-reward success, or a low-reward
failure. A write-all variant records every episode; the comparison is deferred to the
extended version.

\textbf{Manage.} The capacity-bounded buffer evicts by an importance score
combining recency and retrieval frequency (a FIFO variant evicts the oldest);
eviction is deliberately outcome-agnostic, so the outcome signal enters only through
writes and labelling, not retention.

\textbf{Read.} Each arrival queries $M^B$ by a text key over the slice and topology
signature, filtered by violation and strategy tags and ranked by BM25 with recency
weighting, returning a small fixed number of episodes.

\textbf{Use.} Retrieved episodes are injected into Agent~B's prompt as labelled few-shot
examples: positive for plans that survived verification, counter-examples for
those that did not, each carrying the concrete slice, plan shape, and outcome from a
similar topology context. Agent~B thus conditions its next plan on which plan shapes
verifiably survived on comparable substrates rather than on its own prior guesses.

The write rule, the eviction rule, and the recency weighting are the only axes on which
the two memory arms differ; the retrieval key, the tag schema, and the prompt format are
shared.

\subsection{Plan cache}
ORION caches abstract plans using a key composed of the slice type, QoS bucket, and SFC template. A cached plan is revalidated against the current substrate before reuse and invalidated when drift is detected. The cache is disabled in the memory-isolation experiments so that each arrival exercices the retrieval pathway; its effect on LLM invocation cost is evaluated separately. Plan caching is treated as call-economy engineering rather than as a contribution of this work.

\section{RL-based Orchestration}
\label{sec:rl}

The RL layer turns Agent~B's abstract plan into a committed cross-domain
partition. It is a cooperative multi-agent system: an MDO
coordinator that selects the partition under a soft prior from the plan, and
per-domain actors that place VNFs on concrete nodes. The coordinator is a PPO
agent over an enumerated joint-partition space; the actors are graph-attention
networks, architecture-shared without weight-sharing, behaviour-cloned from a
first-fit teacher and held fixed thereafter, so the learned variation across arms is confined to
the coordinator and the plan layer. We describe
the coordinator and its action space, the prior
coupling that carries the plan into it, the domain
actors, the representation choice that the joint
action space embodies, and the training procedure
and reward.

\subsection{MDO coordinator}
\label{sec:rl-mdo}
The coordinator observes an \emph{abstract} view of the substrate: for each
domain $m$ the aggregate residual CPU/RAM $(\hat c^m,\hat r^m)$, its dominant
tier $\tau^m$, and a single-node headroom summary $h^m$; for each inter-domain
link $\ell$ the residual bandwidth $b^{\mathrm{res}}_\ell$ and propagation delay
$D_\ell$; and the plan's per-VNF structure (required tier and resource demands).
Domain features are ordered by the canonical key $(\tau^m, m)$ so the policy is
invariant to domain indexing. The headroom term
$h^m=\max_{n\in m}\min\!\big(\hat c_n/c_{\mathrm{ref}},\,\hat r_n/r_{\mathrm{ref}}\big)$
is the coordinator's only window onto node-level fitness: aggregate residuals
can be ample while no single node is large enough to host a VNF, and $h^m$
exposes exactly that gap.

The suggested partition $\tilde m$ is not part of the observation. It reaches
the coordinator solely through the prior of Section~\ref{sec:rl-prior}; the
observation carries the plan's tier and demand structure but never the domain
assignment. This keeps the two channels distinct: what the plan
\emph{constrains} (tiers, demands, the feasibility mask) enters as observation,
what the plan \emph{recommends} (the split) enters as a prior whose weight is
annealed toward a nonzero floor over training and retained as an inference-time bias.

The policy $\pi^{\mathrm{MDO}}_\phi$ acts over the \emph{joint} partition space.
For a chain of $K$ VNFs on $M$ domains it places a single categorical over the
enumerated set of tier-feasible joint assignments
$\prod_{k=1}^{K} D(\tau_{f_k})\subseteq\{1,\dots,M\}^{K}$, where $D(\tau_{f_k})$
is the set of domains whose tier supports VNF $f_k$; tier-infeasible joint
assignments are masked to zero probability. Tier feasibility is the \emph{only}
hard mask. Aggregate-resource feasibility is deliberately left unmasked and
learned as a soft signal from $h^m$, since a domain may hold ample total
residual yet no admitting node. A shared MLP encoder feeds the joint actor head
and an auxiliary value head $V^{\mathrm{MDO}}_\psi$ that survives at inference to
drive an optional value-based rejection trigger. The joint parameterisation is
the deliberate design choice analysed in Section~\ref{sec:rl-repr}: it is what
lets the policy represent cross-VNF feasibility couplings such as co-location,
which a per-VNF factorisation cannot express.

\subsection{Soft-prior coupling to the plan}
\label{sec:rl-prior}
Agent~B's suggested configuration enters the coordinator's objective as a KL
regulariser over the joint action space, following the value-based view of
LLM-guided RL~\cite{linvit-hu,yan-priors}:
\begin{equation}
\label{eq:mdo-loss}
\mathcal{L}^{\mathrm{MDO}}
  = \mathcal{L}^{\mathrm{PPO}}
  + \beta_t\,\mathrm{KL}\!\big(\pi^{\mathrm{MDO}}_\phi(\cdot\mid o^{\mathrm{MDO}}_t)
        \,\big\|\,\pi_{\mathrm{prior}}(\cdot\mid \tilde m_t)\big),
\end{equation}
where $\pi_{\mathrm{prior}}$ is a temperature-softened joint distribution peaked
on the suggested configuration $\tilde m_t$. Both distributions are discrete
categoricals over the same enumerated joint support, so the KL is evaluated in
closed form and differentiated directly by autograd, which sidesteps by
construction the Monte-Carlo gradient-estimator pitfalls that arise only when the
KL is estimated by sampling. The weight $\beta_t$ is annealed linearly toward a nonzero floor: strong
early, so the plan guides exploration before the value estimate stabilises;
reduced but never eliminated late, so the plan keeps advising even after the
coordinator's own advantage estimates mature. The KL term acts during training;
at inference the suggested partition is retained as a decode-time bias on the
coordinator's action logits, so the LLM continues to advise the committed
partition at decision time. The objective's closed-form optimum reweights the plan prior by the exponentiated
advantage; SayCan~\cite{saycan} is its greedy limit and unregularised RL its opposite
limit, and ORION operates between them at the nonzero floor, with the decode-time
bias carrying the plan into inference.

\subsection{Per-domain actors}
\label{sec:rl-actors}
Each domain runs an actor on a strictly local observation, its own per-node
residuals, intra-domain links, and the plan fragment assigned to it, and returns
a compact response: a feasibility flag $z^m$ and, when feasible, its
intra-domain delay $\delta^m$ and resource cost $\mathrm{cost}^m$. Concrete node
and path selection stays internal to the actor and never crosses the domain
boundary, which is what makes the abstraction gradient consistent \emph{as a
policy-input hierarchy}: the planner's policy conditions on published aggregates,
the coordinator's on aggregates plus $h^m$, and only the actor's on per-node
inventory. The actor is a graph-attention network (GATv2)
over the domain subgraph with residual bandwidth and propagation delay as native
edge features, under architecture-sharing without weight-sharing so every domain
uses the same architecture with its own parameters. Each actor is warm-started by
behaviour cloning on a first-fit teacher, and the warm start passes a gated audit
against the teacher before it is used, so execution starts from a competent,
verified policy rather than from scratch. In all experiments reported here the audited actors are then held \emph{fixed},
confining the learned variation across arms to the planner and the coordinator.

\subsection{Action representation: joint vs.\ factored}
\label{sec:rl-repr}
A \emph{factored} policy that emits an independent per-VNF categorical assumes the
domain assignments of different VNFs are conditionally independent given the
observation. Co-location feasibility (a chain is admissible only if a set of VNFs
land \emph{together}) and cut feasibility (admission depends on \emph{which}
boundary the split falls at) are joint properties of the assignment vector that no
product of per-VNF marginals represents, so a factored policy spreads mass across
jointly infeasible splits, whereas the joint policy of Section~\ref{sec:rl-mdo}
concentrates on configurations that survive verification. The cost is size, the
enumerated support grows as $\mathcal{O}(M^{K})$, which the tier-feasibility mask
prunes to a tractable set at the chain lengths and domain counts here (we report
the resulting sizes with the setup); beyond that regime, feasibility-pruned or
autoregressive decoding that never materialises the full product space is the
mitigation, a scaling direction we do not claim to solve.

\subsection{Training procedure and reward}
\label{sec:rl-train}
Training follows centralised-training / decentralised-execution: a single shared
critic $V_\phi(s_t)$ reads the full joint state during training only and is
discarded at inference, while the coordinator and actors act on their own
observations at both training and inference. Slice requests arrive as a Poisson
process and each is resolved by the single-shot select--place--verify loop of
Section~\ref{sec:arch}, the coordinator sampling one partition under the prior of
Eq.~\eqref{eq:mdo-loss} and the involved actors returning $(z^m,\delta^m,
\mathrm{cost}^m)$. Each arrival yields one transition per
agent, all sharing the global reward $R_t$; advantages are computed by GAE and the
coordinator is updated by the PPO clipped surrogate, its loss
carrying the extra KL term of Eq.~\eqref{eq:mdo-loss}. The domain actors remain at
their audited behaviour-cloned weights throughout
(Section~\ref{sec:rl-actors}); the centralised critic is therefore used to train
the coordinator, and the framework admits joint actor updates without change
should the execution layer become the binding constraint. Outcomes are written to
the episodic memory $M^B$ during training under the selective-write rule of
Section~\ref{sec:llm}; the semantic stores are read-only.

The reward is terminal and shared across the arrival:
\begin{equation}
\label{eq:reward}
\begin{aligned}
R_t = {}&\underbrace{\mu\, z_s}_{\text{admission}}
      \;-\;\underbrace{\alpha\, \mathrm{Cost}(\pi^\ast)}_{\text{efficiency}}\\
      &\;-\;\underbrace{\lambda_{\mathrm{viol}}\,\mathbf{1}[\text{C2/C3/C5b/C7}]}_{\text{hard penalty}}
      \;+\;\underbrace{\eta\, \mathrm{LocalScore}(\pi^\ast)}_{\text{quality shaping}},
\end{aligned}
\end{equation}
with $\mu\!\gg\!\alpha,\eta$ so admission dominates ($\mu\!=\!100$,
$\alpha\!=\!1$, $\lambda_{\mathrm{viol}}\!=\!10$, $\eta\!=\!1$). Here
$z_s\!\in\!\{0,1\}$ is set only if the slice both commits and survives
verification; $\mathrm{Cost}$ aggregates CPU, RAM, and bandwidth with
inter-domain links penalised above intra-domain; and $\mathrm{LocalScore}$ is the
clipped cost improvement over a greedy first-fit placement of the same request.
The hard penalty fires from the verifier's post-commit verdict under load-dependent
queueing, not from the light-load pre-commit estimate. The implementation also carries a
retry-penalty term $-\xi\,(T_t-1)$ for the multi-trial partition loop; the
evaluated configuration resolves each arrival in a single shot ($T_t\!=\!1$), so
this term is inert and omitted from Eq.~\eqref{eq:reward}.

\section{Experimental Results}
\label{sec:exp}

This section reports what is measured and states the protocol for what is not yet.
We first ground the LLM stages with the measured call economy
(Section~\ref{sec:exp-cost}), then give the paper's primary evidence for the
memory claim: a plan-quality isolation on held-out topology families in which the
selection and execution layers are held fixed, so the plan is the only varying
factor (Section~\ref{sec:exp-prelim}). The full stack couples that plan into the
trained MDO coordinator, which acts with the frozen behaviour-cloned domain actors
of Section~\ref{sec:rl-actors}; its evaluation follows the pre-specified protocol of
Section~\ref{sec:exp-protocol}, with results in the extended version. The diagnostic
characterization that motivates this design is Section~\ref{sec:diagnosis}.

\subsection{Setup}
\label{sec:exp-setup}

\textbf{Scenario classes.} Two workload regimes. The \emph{conventional} class fixes
the per-VNF volume-change ratio to $\mathrm{VCR}=1.0$ and holds inter-VNF bandwidth
constant along the chain, the setting most prior slicing work assumes; the
\emph{stress} class amplifies the VCR so bandwidth grows along the chain, tightening
the coupling between placement and inter-domain routing. Reporting both separates
behaviour on the conventional regime from behaviour when bandwidth structure is the
binding constraint.

\textbf{Topology families and split.} We use the parametrized generator of
Section~\ref{sec:diagnosis}, crossing domain capacity, tier coverage, and
inter-domain bandwidth. Training uses four families; evaluation is on four
\emph{held-out} families never seen in training, so every system-level result is
a transfer claim across topology signatures.

\textbf{Topology signature.} Retrieval and outcome labelling are keyed on a
\emph{topology signature}: the three generator axes of Section~\ref{sec:diagnosis}
quantized, with the slice type and chain template. Two arrivals share a signature
when they face structurally equivalent substrates, which is what lets an episode
observed on one instance inform planning on another.

\textbf{Arms.} The isolation of Section~\ref{sec:exp-prelim} varies only the plan
layer, holding selection and execution fixed; the six arms of the full-stack
protocol, and the ladder they form, are defined in Section~\ref{sec:exp-protocol}.

\textbf{Reproducibility.} Every number is produced by a run whose code commit and
configuration are recorded, and no result is quoted before its provenance verifies.

\subsection{LLM call economy}
\label{sec:exp-cost}
The two LLM stages are the dominant per-decision cost, and the architecture keeps
them off the hot path by design. On the profiled deployment run
(13{,}204 LLM calls over 32{,}435 RL decisions), a single LLM call takes a median
$3.94$~s and consumes a median $0.98$~kJ of GPU energy, against $5.0$~ms and
$0.56$~J for one RL selection decision, a gap of roughly $790\times$ in time and
$1750\times$ in energy per decision. The LLM accounts for $99.9\%$ of
decision-path latency; retrieval ($1.29$~ms), memory write ($3.87$~ms), and
routing ($2.20$~ms) are negligible beside it. A representative $20$-hour run
draws $4.26$~kWh end to end. The design implication is direct: invoke the LLM
rarely (cache-gated, on signature miss or drift) and keep every downstream
component cheap.

These figures make cache-gated invocation an admission-control property, not only
an energy one: with a mean LLM call at $3.94$~s against a $5$~ms selection path, a
synchronously resolving orchestrator is call-rate bound until plan reuse drives the
LLM off the per-arrival path. 

\subsection{Plan-quality isolation with a fixed selector}
\label{sec:exp-prelim}
Coupling the plan into a trained selector varies two things at once; to show the
memory effect is not an artifact of the selector, we isolate it here with a
five-arm study that varies plan quality alone, holding the selection and execution
layers fixed at a deterministic prior-following path. Here every arm commits the
plan's suggested partition unchanged, so any admission difference is attributable to
the plan alone. Its system-level validation, once the plan is coupled into a trained
coordinator, follows the pre-specified protocol of Section~\ref{sec:exp-protocol}.
\begin{table}[t]
  \centering
  \caption{Plan-quality isolation on held-out topology families: admission as \%
  of the verifier ceiling with the selection and execution layers held fixed at a
  deterministic prior-following path, so the plan is the only varying factor.
  \textbf{The two upper rows are information-privileged references, not competing
  arms}: both read the full per-node substrate, which no arm in the lower block
  receives. The meaningful comparisons are
  \emph{within} the abstract-view block.}
  \label{tab:exp-prelim}
  \small
  \setlength{\tabcolsep}{6pt}
  \begin{tabular}{@{}lcc@{}}
    \toprule
    Arm & Seed 0 & Seed 1 \\
    \midrule
    \multicolumn{3}{@{}l}{\emph{Full-substrate references (information-privileged)}}\\
    RA-ColocFB                    & 94.8 & 93.6 \\
    Plain-ColocFB                 & 94.7 & 90.2 \\
    \midrule
    \multicolumn{3}{@{}l}{\emph{Abstract per-domain view (comparison block)}}\\
    Full-$M^B$                    & \textbf{61.5} & \textbf{65.8} \\
    Memory-off                    & 56.9 & 54.3 \\
    FIFO-$M^B$                    & 50.4 & 53.5 \\
    \bottomrule
  \end{tabular}
\end{table}

Table~\ref{tab:exp-prelim} reports the isolation. Both reference rows are static
co-location-first heuristics with first-fit fallback: Plain-ColocFB applies the rule
directly, while RA-ColocFB is routability-aware, preferring splits whose
inter-domain flows it can route. The two ColocFB rows are separated
deliberately: both operate with full per-node visibility of every domain, which no
LLM arm receives and which no deployable multi-domain orchestrator would have, since
domains publish aggregate state across the management boundary
(Section~\ref{sec:model}). They bound what is attainable with information the
learned arms are denied by construction, and the comparisons that carry the claim
are within the abstract-view block.
Full-$M^B$ admits more than the
memory-off control on both seeds ($+4.6$ and $+11.5$ points), and more than the
FIFO-$M^B$ ablation on both ($+11.1$ and $+12.3$ points). With the selector fixed,
the episodic memory therefore improves plan quality on topology families never seen
during warm-up, and the gap to FIFO-$M^B$ locates the advantage in \emph{which}
episodes are retained and retrieved rather than in the presence of a retrieval
mechanism: an identical store under first-in-first-out admission loses most of the
benefit. The sign is consistent across both seeds; the magnitude is not established.

\subsection{Evaluation protocol for the full stack}
\label{sec:exp-protocol}
The isolation above fixes the selection and execution layers. Coupling the plan into
the trained coordinator is evaluated under a protocol fixed in advance of the runs,
stated here so the analysis is pre-specified rather than selected after the fact. Each arm is run on both
scenario classes over the held-out family split, scored against the verifier ceiling,
with admission as \% of that ceiling reported per seed; each arm runs at least three
instance seeds of $100$ arrivals per held-out family, a floor fixed in advance
because single-seed differences on these substrates fall within seed variance. \textbf{Arms.} All learned arms share the trained coordinator and the frozen
behaviour-cloned actors, differing only in what reaches Agent~B's prompt and how the
episodic store is labelled, so any difference is attributable to the plan layer.
\emph{Full ORION} conditions on the slice request, the abstract topology, $K^B$
entries and outcome-labelled $M^B$ episodes; \emph{Memory-off} drops $M^B$;
\emph{Reward-labelled memory} keeps Full's store, retrieval and prompt slots but
labels episodes by the scalar admission reward instead of the feasibility verdict
(the ARC-style regime of Table~\ref{tab:related}); \emph{Prior-only} makes no
inference-time LLM call: a co-location-first plan-builder supplies the suggested
partition, which enters only through the observation and the tier-feasibility mask,
and the coordinator commits its own argmax. The decode-time bias of
Section~\ref{sec:rl-prior} is applied only by the arms that call the LLM at
inference, so neither arm here receives it. \emph{RL-alone} is identical at
inference but trained with $\beta_t=0$, so the two differ only in training. \emph{Plain} is the
co-location-first heuristic with first-fit fallback and, like the ColocFB references
of Section~\ref{sec:exp-prelim}, reads the full per-node substrate.

The protocol asks five
questions, each a single contrast between two arms that differ in exactly one
component:
\begin{enumerate}
\item \textbf{Training prior.} Does training under the persistent KL prior beat the
same stack trained with $\beta_t=0$? The two arms are identical at inference, both
committing their own argmax partition over a heuristic-supplied plan, so the
contrast isolates training. (Prior-only vs.\ RL-alone.)
\item \textbf{Inference-time LLM channel.} Does the LLM advising at decision time,
as a live plan grounded by $K^B$ together with its decode-time bias, beat a
heuristic-supplied plan committed by the same trained selector? This contrast varies
the whole inference-time channel rather than the plan alone. (Memory-off vs.\
Prior-only.)
\item \textbf{Episodic memory.} Does $M^B$ improve admission over the identical
trained stack with the memory disabled? (Full vs.\ Memory-off.)
\item \textbf{Memory signal.} Holding store, retrieval, and prompt slots fixed, does
the feasibility verdict beat the scalar admission reward as the labelling and gating
signal? (Full vs.\ Reward-labelled memory.) This is the central test: a positive gap
attributes the advantage to the \emph{signal} rather than to the presence of a
retrieval mechanism.
\item \textbf{System vs.\ heuristic.} Does the learned stack admit more than the
co-location-first heuristic, and on which scenario class does any advantage
concentrate? (Full vs.\ Plain, reading Plain as the information-privileged reference
of Section~\ref{sec:exp-prelim}.)
\end{enumerate}
Signs are reported as measured, including where they run against the hypothesis.

\subsection{A pre-specified boundary probe}
\label{sec:exp-boundary}
An outcome-driven memory helps only where a better plan shape exists to be recalled.
On routing-critical families, where admission turns on cross-domain cut points and
inter-domain paths rather than on tier assignment, the memory's advantage may
invert, because the recalled plan shape does not transfer when the binding structure
is the route, not the placement. We therefore pre-specify this as a boundary probe
rather than an aggregate: the comparison is reported per family and scoped to the
stress class, so a family where the mechanism fails is identified rather than
averaged away.

\section{Conclusion}
\label{sec:conclusion}
In this paper, we presented ORION, a hierarchical orchestrator for multi-domain 6G slice placement
whose LLM planner adapts its context from verifier-checked feasibility outcomes associated with concrete plan
structures and network conditions. Our diagnostic study showed that rejected requests are predominantly feasible but remain undiscovered, that fixed plan-building strategies do not perform consistently across topology regimes, and that improving the partition selector alone provides limited gains under sequential arrivals.
In a controlled plan-quality evaluation on held-out topology families, the outcome-driven memory improved admission by $4.6$ to $11.5$ percentage points over
the same planner without memory, while the full memory configuration consistently outperformed its FIFO variants. The latency measurements further show that LLM invocation dominates the decision time, motivating cache-gated plan reuse. A full-stack evaluation, including the reward-labelled-memory control and routing critical boundary analysis, is reserved for future work.

\bibliographystyle{IEEEtran}
\bibliography{orion-paper}

\end{document}